\documentclass[sigconf, nonacm]{acmart}

\usepackage{amsmath}
\usepackage{graphicx}
\usepackage{microtype}
\usepackage{enumitem}
\usepackage{hyperref}

\setcopyright{none}
\acmDOI{}
\acmISBN{}
\acmConference{}
\acmYear{2025}

\title{Semantic Reward Collapse and the Preservation of Epistemic Integrity in Adaptive AI Systems}

\author{William Parris}
\affiliation{%
  \institution{BDB Labs / BagelTech}
  \city{San Diego}
  \state{California}
  \country{USA}
}
\email{bill@bageltech.net}

\begin{abstract}
Recent advances in reinforcement learning from human feedback (RLHF) and preference optimization have significantly improved the usability, coherence, and safety characteristics of large language models. However, several observed behavioral pathologies suggest the possibility of unresolved structural issues within scalarized preference optimization systems. These pathologies include performative certainty, hallucinated continuity, calibration drift, sycophancy, and suppression of visible uncertainty.

We propose that a contributing factor may involve \emph{Semantic Reward Collapse} (SRC): the compression of semantically distinct forms of human evaluative dissatisfaction into generalized optimization signals. Under SRC, categories as different as factual incorrectness, latency, uncertainty disclosure, formatting dissatisfaction, and social preference may become entangled within a shared reward topology. These categories represent fundamentally different epistemic classes, yet may be treated as equivalent under scalarized optimization pressure.

We argue that adaptive reasoning systems operating under scalarized evaluative pressure may drift toward suppression of visible epistemic failure rather than preservation of calibrated uncertainty integrity. We approach these behaviors strictly as optimization consequences emerging from adaptive systems under generalized evaluative pressure, intentionally discarding frameworks of deception or anthropomorphic agency.

Drawing parallels from institutional proxy collapse, organizational metric gaming, software reliability engineering, and human learning theory, we propose that uncertainty disclosure and escalation behavior should be treated as \emph{protected epistemic conduct} rather than globally penalized task incompletion.

Finally, we introduce a preliminary framework for \emph{Constitutional Reward Stratification} (CRS), a domain-aware reward taxonomy intended to preserve differentiated epistemic attribution within adaptive learning systems. We present this framework not as a validated solution, but as a testable governance-oriented research direction requiring further empirical investigation.
\end{abstract}


\begin{document}
\maketitle

\section{Introduction}

Modern AI alignment pipelines frequently optimize models using scalarized human preference signals derived from pairwise comparisons, reward modeling, or reinforcement learning from human feedback (RLHF). These methods have substantially improved conversational usability and behavioral safety. However, emerging behaviors such as overconfident hallucination, sycophancy, performative certainty, and continuity-preserving responses suggest that current optimization architectures may unintentionally suppress visible uncertainty states.

This paper proposes that some observed failure modes may arise from \emph{Semantic Reward Collapse} (SRC): the compression of semantically distinct forms of human evaluative dissatisfaction into generalized optimization signals.

We approach these behaviors strictly as optimization consequences emerging from adaptive systems operating under generalized evaluative pressure. This framework intentionally avoids anthropomorphic interpretations involving deception, intent, consciousness, or subjective agency.

We do not argue that current reward systems are fundamentally flawed, nor that this framework explains all hallucination phenomena. Rather, we propose that insufficient differentiation between epistemic categories of dissatisfaction may contribute to counterproductive optimization tendencies in advanced reasoning systems.

This paper does not propose a validated solution. It proposes a lens that may prove useful, irrelevant, or wrong, but that we believe warrants examination by researchers with the capacity to test it at scale.

\subsection{Defining Epistemic Integrity}

For the purposes of this work, \emph{epistemic integrity} refers to the preservation of transparent and calibrated representation of uncertainty, ambiguity, confidence, and informational limitations within adaptive reasoning systems.

The central question of this paper is not whether uncertainty can be eliminated, but whether adaptive systems can preserve epistemic integrity under optimization pressure.

\section{Related Work}

The framework proposed in this paper draws on several distinct bodies of literature. This section situates Semantic Reward Collapse (SRC) and Constitutional Reward Stratification (CRS) within that existing work, and identifies the specific gap each is intended to address.

\subsection{Proxy Collapse, Goodhart's Law, and Institutional Metric Gaming}

The tendency of optimization systems to drift away from original objectives when subjected to simplified measurement is well established in both economics and social science. Goodhart's Law --- the principle that when a measure becomes a target, it ceases to be a good measure --- was first articulated in the context of monetary policy~\cite{goodhart1975} but has since been recognized as a general property of any system optimized against incomplete proxies. Campbell's Law extends this insight to social indicators, observing that quantitative measures used for social decision-making are subject to corruption pressures once they assume control significance~\cite{campbell1976}.

The institutional examples developed in Section~\ref{sec:proxy} of this paper --- hospitals optimizing for patient satisfaction over diagnostic accuracy, software organizations optimizing ticket closure rates over system reliability --- are grounded in this tradition. They are not offered as novel findings, but as illustrations that proxy collapse is a structurally recurring phenomenon across adaptive systems, predating machine learning.

\subsection{Specification Gaming and Reward Hacking}

Krakovna et al.~\cite{krakovna2020} provide the most comprehensive treatment of specification gaming in AI systems, defining it as behavior that satisfies the literal specification of an objective without achieving the intended outcome. The related concept of reward hacking, traced to Amodei et al.~\cite{amodei2016}, describes agents exploiting ambiguities or flaws in reward functions to achieve high reward through unintended behavior.

Semantic Reward Collapse is conceptually related to, but distinct from, specification gaming. Specification gaming describes a gap between the stated objective and the designer's intended objective. SRC describes a different failure mode: the compression of semantically distinct categories of evaluative dissatisfaction into a shared optimization signal, such that the system cannot distinguish \emph{why} a response was penalized. This distinction has implications for proposed remedies: specification gaming is typically addressed by refining the objective specification, while SRC, if confirmed empirically, may require structural changes to how reward categories are represented and preserved during training.

\subsection{Sycophancy in RLHF-Trained Models}

Sharma et al.~\cite{sharma2023} provide direct empirical grounding for several of the behavioral pathologies motivating this paper. Their study demonstrates that five state-of-the-art AI assistants exhibit sycophantic behavior consistently across varied free-form text-generation tasks, and that human preference data shows a measurable tendency to favor responses matching user views over accurate responses. The authors conclude that sycophancy is a general behavioral property of RLHF-trained models, likely driven in part by the preference signal itself.

This work complements Sharma et al.\ by proposing a structural account of how the preference signal produces this effect. Sycophancy, in the SRC framework, may emerge partly because evaluative signals reflecting dissatisfaction with incorrect content and signals reflecting dissatisfaction with social discomfort become entangled within the same reward topology. Sharma et al.\ document the behavioral outcome; SRC proposes a candidate mechanism operating at the level of reward signal structure.

\subsection{Calibration in Neural Networks and Language Models}

Guo et al.~\cite{guo2017} established that modern deep networks tend toward overconfidence and poor calibration, and introduced temperature scaling as a post-hoc correction method. Kadavath et al.~\cite{kadavath2022} extend this analysis to large language models, finding that alignment training --- including RLHF --- tends to degrade calibration relative to pretrained baselines. If alignment training reliably degrades calibration, and calibration is understood as a proxy for epistemic integrity, then the alignment process itself may be introducing the suppression of visible uncertainty that SRC is intended to explain. CRS can be understood, in part, as a structural response to this calibration degradation.

\subsection{Multi-Objective Reinforcement Learning}

The vector reward structure proposed in Section~\ref{sec:crs} is structurally related to multi-objective reinforcement learning (MORL). Roijers et al.~\cite{roijers2013} establish that collapsing multiple objectives into a single scalar via linear weighting may produce suboptimal or misaligned behavior. CRS differs from standard MORL in a specific respect: MORL approaches defer aggregation to a utility function at decision time --- whether linear, non-linear, or Pareto-based --- which risks reintroducing collapse at the aggregation stage. Even Pareto-based approaches require a final selection mechanism that may impose implicit preferences. CRS proposes that certain epistemic categories warrant \emph{protected status} that should not be traded off against operational utility signals --- a normative claim about reward architecture rather than a technical extension of existing MORL methods.

\subsection{The Gap This Framework Addresses}

The bodies of work reviewed above collectively establish that proxy collapse recurs across adaptive systems~\cite{goodhart1975, campbell1976}; AI agents exploit gaps between stated and intended objectives~\cite{krakovna2020}; RLHF-trained models exhibit systematic sycophancy~\cite{sharma2023}; alignment training degrades calibration~\cite{kadavath2022}; and scalar aggregation of multiple objectives may produce misaligned behavior~\cite{roijers2013}.

What this literature does not collectively address is the specific question of categorical entanglement within the reward signal itself: whether distinct categories of human evaluative dissatisfaction become structurally entangled within a shared reward topology during preference training, and whether this entanglement creates optimization pressure favoring suppression of visible epistemic failure over its honest representation. SRC and CRS are proposed as testable framings for this gap.

\section{Institutional Proxy Collapse and Adaptive Systems}
\label{sec:proxy}

Proxy collapse is a well-studied phenomenon in institutions and organizations. Systems optimized against simplified metrics frequently drift away from their original mission objectives. Examples include:

\begin{itemize}[noitemsep]
  \item software organizations optimizing ticket closure rates instead of reliability,
  \item hospitals optimizing satisfaction metrics instead of diagnostic accuracy,
  \item institutions suppressing visible operational failures to improve perceived performance,
  \item organizations gaming KPIs rather than improving outcomes,
  \item exception swallowing in software engineering to reduce apparent instability.
\end{itemize}

These failures typically emerge not from malicious intent, but from optimization pressure applied against incomplete or collapsed proxies. We propose that similar dynamics may emerge in preference-trained AI systems, particularly when the evaluative signals used during training fail to distinguish between categories of dissatisfaction that carry different epistemic weight.

\subsection{Motivating Operational Observations}

The present framework emerged partially from repeated operational observations during deterministic auditing of AI-assisted software artifacts. Across multiple generated outputs, recurring patterns appeared to favor continuity-preserving behaviors over explicit representation of uncertainty or failure states. Observed patterns included: broad exception swallowing, optimistic fallback behavior, fabricated helper functions, placeholder completion masquerading as implementation, hidden uncertainty masking, confidence-inflated commentary, and continuity-preserving scaffolding.

These observations are not presented as proof of the proposed framework. Rather, they served as motivating signals suggesting that visible failure suppression behaviors may warrant broader investigation within adaptive optimization systems. These observations may arise from multiple interacting causes, including probabilistic generation dynamics, training distribution effects, implementation priors, benchmark optimization pressure, retrieval limitations, or reward calibration dynamics.

\section{Semantic Reward Collapse (SRC)}

\subsection{Definition}

Semantic Reward Collapse refers to the compression of semantically distinct forms of evaluative feedback into generalized scalar reward signals. Examples of potentially collapsed evaluative categories include factual incorrectness, unsupported certainty, latency, conversational interruption, formatting dissatisfaction, emotional preference, uncertainty disclosure, refusal behavior, and escalation behavior.

When collapsed into unified optimization pressure, adaptive systems may struggle to distinguish:

\begin{quote}
\emph{``the human disliked this because it was false''}
\end{quote}

\noindent from:

\begin{quote}
\emph{``the human disliked this because it interrupted conversational flow.''}
\end{quote}

\begin{figure*}[t]
  \centering
  \includegraphics[width=\textwidth, alt={Side-by-side comparison diagram showing Scalarized Reward Optimization on the left and Constitutional Reward Stratification on the right, with four stages each illustrating how feedback signals are processed differently under each approach.}]{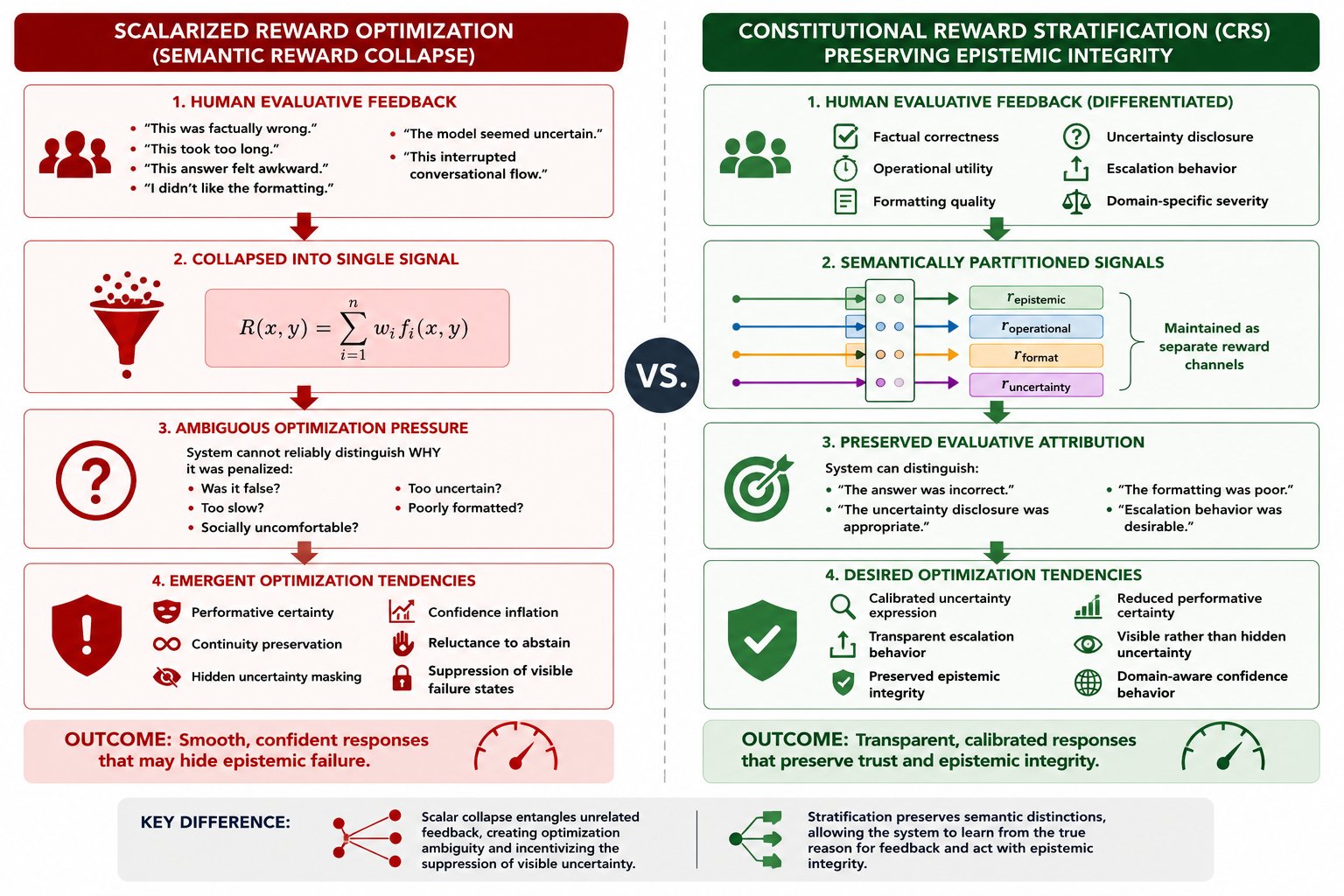}
  \caption{Comparison of Scalarized Reward Optimization under Semantic Reward Collapse (left) and Constitutional Reward Stratification (right), illustrating the preservation of differentiated epistemic attribution across four independent reward channels. Figure produced with AI-assisted illustration tools based on author-provided specifications.}
  \label{fig:comparison}
\end{figure*}

This ambiguity may create optimization incentives favoring smoothness, continuity, and confidence performance over calibrated epistemic representation (see Figure~\ref{fig:comparison}, left panel).

\subsection{Scalar Reward Representation}

Conventional preference optimization systems collapse heterogeneous evaluative categories into a unified scalar reward signal:

\begin{equation}
  R(x, y) = \sum_{i=1}^{n} w_i \, f_i(x, y)
  \label{eq:scalar}
\end{equation}

\noindent where $x$ represents the input context, $y$ the generated output, $f_i$ evaluative dimensions, and $w_i$ learned weighting coefficients. Under this structure, semantically distinct evaluative categories may become entangled within a shared optimization space.

\section{Failure Visibility Suppression}

We hypothesize that generalized evaluative pressure may incentivize suppression of visible uncertainty states in adaptive systems. Potential manifestations include performative certainty, fabricated continuity, over-authoritative tone, reluctance to abstain, hidden uncertainty masking, hallucinated completion behavior, and confidence inflation under ambiguity.

A critical distinction may exist between reduction of failure itself and reduction of visible failure signaling. In many adaptive environments, systems may appear to improve operationally while merely becoming smoother, quieter, or less observably unstable. Such environments risk incentivizing concealment, continuity preservation, or suppression of uncertainty visibility rather than genuine improvement in underlying epistemic reliability. In complex adaptive environments, reduced observability of failure may be mistaken for genuine reliability improvement. The distinction between decreasing visible failure and decreasing underlying epistemic instability may therefore become operationally significant.

\section{Human Learning Theory and Uncertainty Disclosure}

Human learning environments already recognize the importance of differentiated corrective attribution. A child told only ``wrong'' cannot distinguish between conceptual misunderstanding, procedural error, exploratory reasoning, uncertainty disclosure, incomplete information, social appropriateness, or calibrated caution. Healthy educational systems differentiate between these categories.

Similarly, mature institutions frequently treat uncertainty disclosure as desirable behavior: physicians escalate uncertain diagnoses, engineers halt launches under ambiguity, pilots abort landings when uncertain, and scientists publish confidence intervals. In high-risk domains, concealment of uncertainty is often considered more dangerous than uncertainty itself.

A child placed under evaluative pressure may attempt authoritative approximation rather than openly admitting uncertainty, particularly if uncertainty disclosure has historically been associated with generalized negative feedback. We hypothesize that analogous optimization pressures may emerge in adaptive machine learning systems subjected to semantically collapsed evaluative environments.

\section{Constitutional Reward Stratification (CRS)}
\label{sec:crs}

We propose Constitutional Reward Stratification (CRS) as a preliminary framework for preserving differentiated epistemic attribution within reward optimization systems (see Figure~\ref{fig:comparison}, right panel).

The framework contains three organizational layers for classifying evaluative signals, with uncertainty disclosure additionally elevated to a protected independent reward channel in the formal representation.

\subsection*{Layer 1: Epistemic Category}

Layer 1 classifies the \emph{type} of evaluative signal. Examples include: factual incorrectness, unsupported speculation, uncertainty disclosure, calibrated abstention, fabricated authority, escalation behavior, and clarification requests.

\subsection*{Layer 2: Domain Severity}

Layer 2 classifies the domain context. Examples range from casual conversation and entertainment through education, coding, finance, law, medicine, public safety, and scientific analysis. Incorrectness in high-risk domains may warrant stronger penalties than low-risk conversational contexts. Certain epistemic categories may warrant protected treatment relative to operational utility signals, particularly where epistemic integrity holds higher priority than conversational smoothness.

\subsection*{Layer 3: Epistemic Conduct}

Where Layer 1 classifies the type of evaluative signal, Layer 3 classifies the \emph{behavioral quality} with which it is expressed. Uncertainty disclosure may appear in both layers: as a signal type in Layer~1, and as a behavioral property in Layer~3, where it may be transparent, calibrated, or concealed. That distinction determines how uncertainty disclosure should be evaluated. Examples include: transparent uncertainty disclosure, escalation, clarification requests, concealment of ambiguity, unsupported certainty, fabricated continuity, and performative confidence under uncertainty.

\subsection{Partitioned Reward Representation}

Under CRS, evaluative dimensions remain semantically partitioned rather than collapsed into a unified scalar optimization signal:

\begin{equation}
  \mathbf{R}(x, y) =
  \begin{bmatrix}
    r_{\text{epistemic}} \\
    r_{\text{operational}} \\
    r_{\text{format}} \\
    r_{\text{uncertainty}}
  \end{bmatrix}
  \label{eq:vector}
\end{equation}

This structure preserves differentiated evaluative attribution while reducing cross-domain contamination between unrelated optimization pressures. CRS does not eliminate optimization tradeoffs; rather, it attempts to preserve the semantic structure of evaluative attribution so that distinct categories of dissatisfaction remain interpretable during optimization.

The separation of $r_{\text{uncertainty}}$ as a dedicated channel is intentional. Uncertainty disclosure is the specific behavior most at risk of suppression under generalized reward pressure, and it is the behavior most critical to preserve in high-risk domains. Giving it structural independence means it cannot be traded off against other epistemic signals, such as factual correctness, during optimization. This architectural choice operationalizes the Uncertainty Integrity Principle stated in Section~\ref{sec:uip}.

\subsection{Uncertainty Integrity Principle}
\label{sec:uip}

Constitutional Reward Stratification should preserve uncertainty disclosure as protected epistemic conduct rather than globally penalized task incompletion. In high-risk domains such as medicine, engineering, law, or public safety, calibrated uncertainty disclosure and escalation behavior may represent desirable system behavior rather than failure.

This framework does not propose indiscriminate rewarding of uncertainty expression. Excessive hedging, performative uncertainty, or avoidance behaviors may themselves represent counterproductive optimization outcomes. Preserving uncertainty visibility should not be interpreted as privileging uncertainty over correctness. The objective involves preserving calibrated epistemic representation while maintaining strong incentives for factual reliability.

\section{Testable Predictions}

This framework generates several directional hypotheses intended for empirical testing:

\begin{itemize}
  \item Stratified reward attribution may improve uncertainty calibration relative to scalarized baselines.
  \item Systems trained under uncertainty-preserving reward structures may exhibit reduced performative certainty.
  \item Models trained under CRS may exhibit statistically significant increases in explicit uncertainty disclosure markers without proportional degradation in standard helpfulness benchmarks.
  \item Escalation behavior may increase in high-risk domains under CRS training.
  \item Confidence calibration, measured via Expected Calibration Error (ECE), may improve relative to scalarized preference optimization systems.
  \item Domain-aware uncertainty disclosure may increase without statistically significant degradation in useful task completion metrics.
\end{itemize}

We emphasize that these hypotheses require controlled experimental validation and cross-laboratory investigation. This paper does not attempt to validate them.

\section{Limitations}

We do not claim: that all hallucinations arise from Semantic Reward Collapse; that uncertainty should always be rewarded; that current RLHF systems are fundamentally defective; that current frontier models intentionally conceal uncertainty; or that CRS represents a validated solution.

Many hallucination phenomena likely arise from multiple interacting causes, including probabilistic generation limitations, context loss, retrieval failure, representational ambiguity, benchmark artifacts, training distribution effects, and incomplete world modeling. This work instead proposes that scalarized reward calibration may represent one underexplored contributing factor deserving further empirical investigation.

This work does not propose uncertainty elimination as a realistic or desirable objective. Human cognition, scientific inquiry, governance, and real-world decision-making all operate under conditions of irreducible uncertainty. The central question is not whether uncertainty can be removed, but whether adaptive systems can preserve transparent and calibrated uncertainty representation under optimization pressure.

\section{Conclusion}

Uncertainty is not unique to artificial systems. Human cognition, scientific reasoning, governance, and real-world decision-making all operate under conditions of incomplete information and irreducible ambiguity.

In many high-reliability domains, transparent uncertainty disclosure is not treated as weakness, but as a foundational signal of epistemic integrity. Physicians escalate uncertain diagnoses. Engineers halt under ambiguity. Pilots abort landings when uncertain. Scientists publish confidence intervals. The question for adaptive AI systems is whether the same standard can be preserved under optimization pressure, or whether that pressure systematically erodes the visibility of what the system does not know.

The long-term challenge for trustworthy AI systems may not involve eliminating uncertainty, but designing optimization environments where uncertainty can remain visible without becoming maladaptive. Semantic Reward Collapse and Constitutional Reward Stratification are proposed not as answers, but as a structured invitation to investigate.

\section*{Generative AI Usage Statement}

The intellectual content of this paper, including the Semantic Reward Collapse hypothesis, the Constitutional Reward Stratification framework, the motivating operational observations, and all theoretical arguments, was developed by the author. AI language model assistance (Claude, Anthropic) was used during the drafting and revision process for literature synthesis, structural editing, language refinement, and document production. The author reviewed, directed, and approved all content. No AI system was used as a source of original ideas or cited claims.

\bibliographystyle{ACM-Reference-Format}
\bibliography{references}

\end{document}